\documentclass{./styles/SVproc}
\usepackage{epsfig}         % for postscript graphics files
\usepackage{graphics}       % for pdf, bitmapped graphics files
\usepackage{graphicx}
\usepackage{epstopdf}

\usepackage{url}
\usepackage{orcidlink}

\usepackage[dvipsnames]{xcolor}
\usepackage{amsmath} 
\usepackage{amsfonts}

\begin{document}
\mainmatter              % start of a contribution

\title{PRISM: Personalized Refinement of Imitation Skills for Manipulation via Human Instructions}
\titlerunning{RL Refinement of Imitation Policies from Instructions}  % abbreviated title (for running head)
\author{Arnau Boix-Granell \orcidlink{0009-0009-4404-9141}  \and Alberto San-Miguel \orcidlink{0000-0001-5547-2564} \and \\ Magí Dalmau-Moreno \orcidlink{0000-0003-3119-7347} \and Néstor García\orcidlink{0000-0003-3782-1745}}

%Alberto -
%Magí - 0000-0003-3119-7347
%Arnau - 0009-0009-4404-9141
% Néstor -

%
\authorrunning{Arnau Boix-Granell et al.} % abbreviated author list (for running head)
\tocauthor{Arnau Boix-Granell, Alberto San-Miguel-Tello, Magí Dalmau-Moreno, and Néstor García} 
\institute{Eurecat, Centre Tecnològic Catalunya, \\ Robotics \& Automation Unit, Barcelona, Spain\\
\email{\{arnau.boix, alberto.sanmiguel,  magi.dalmau, nestor.garcia\}@eurecat.org}}

\maketitle              % typeset the title of the contribution

%useful for reviewing 
\newcommand{\alberto}[1]{{\color{magenta}{\textbf{Remark Alberto\c{c}:} #1}}}
\newcommand{\arnau}[1]{{\color{red}{\textbf{Remark Arnau:} #1}}}
\newcommand{\magi}[1]{{\color{blue}{\textbf{Remark Magi:} #1}}}
\newcommand{\nestor}[1]{{\color{teal}{\textbf{Remark Néstor:} #1}}}

\begin{abstract}
This paper presents PRISM: an instruction-conditioned refinement method for imitation policies in robotic manipulation. This approach bridges Imitation Learning (IL) and Reinforcement Learning (RL) frameworks into a seamless pipeline, such that an imitation policy on a broad generic task, generated from a set of user-guided demonstrations, can be refined through reinforcement to generate new unseen fine-grain behaviours. The refinement process follows the Eureka~\cite{ma_eureka_2023_2} paradigm, where reward functions for RL are iteratively generated from an initial natural-language task description. Presented approach, builds on top of this mechanism to adapt a refined IL policy of a generic task to new goal configurations and the introduction of constraints by adding also human feedback correction on intermediate rollouts, enabling policy reusability and therefore data efficiency. Results for a pick-and-place task in a simulated scenario show that proposed method outperforms policies without human feedback, improving robustness on deployment and reducing computational burden. \footnote{PRISM Overview Video: \url{https://www.youtube.com/watch?v=yUUkD1LsCWY} \label{footnotevideo}}

%around adapts policies to shifts in goal configuration and constraints, on top of customary phenomena that can be addressed by RL, e.g. changes on dynamic behaviours of objects. Empirical evaluation across a \textcolor{red}{suite} of manipulation tasks demonstrates that instruction-guided refinement reduces RL sample complexity relative to learning from scratch, improves robustness compared with imitation alone, and facilitates efficient adaptation to distributional shifts. The paper further analyzes the roles of automated and human corrective prompts and documents principal failure modes and design recommendations.

% Results show that presented approach reduces refinement steps with respect to RL and IL alone, enabling its usage 

% We would like to encourage you to list your keywords within
% the abstract section using the \keywords{...} command.
\end{abstract}

% \begin{eqnarray*}
% \dot{x}&=&JH' (t,x)\\
% x(0) &=& x(T)
% \end{eqnarray*}

%
\section{Introduction}

Robotic manipulation in unstructured settings demands controllers that are both data-efficient and robust to shifts in dynamics, goals, and constraints. Pure Imitation Learning (IL) can rapidly acquire competent behaviors from a modest set of demonstrations, but is brittle to out-of-distribution events and lacks recovery strategies~\cite{codevilla_end_2018}. Conversely, Reinforcement Learning (RL) can discover robust, reactive behaviors through exploration, yet training RL from scratch is often sample-inefficient and impractical without large-scale interaction or careful reward engineering \cite{haarnoja_soft_2018}. Prior work has therefore proposed hybrid pipelines that combine the sample efficiency of IL with the adaptivity of RL, enabling faster and safer adaptation through refinement techniques \cite{torne_reconciling_2024_2,ankile_imitation_2024_2,yuan_policy_2025_2} 

While IL-based policies offer efficient task initialization, personalization remains crucial for ensuring effective and acceptable human-robot interactions in manipulation settings, especially when involving non-expert users. Variations in user intent, physical capabilities, and subjective comfort -such as preferred grasp strategies, motion speed, or sensitivity to contact forces- can strongly affect the success and safety of collaborative tasks like handovers or constrained transport. Generic policies trained via IL often fail to accommodate these individualized needs, leading to brittle behaviors that degrade trust and usability. Personalization enables robots to adapt to human-specific requirements, improving both functional success and user satisfaction. Studies in assistive and collaborative manipulation show that personalization reduces user workload, increases perceived fluency, and fosters greater trust and willingness to delegate tasks~\cite{canal_personalization_2016_2,maroto-gomez_adaptive_2023_2}. Without such adaptation, even technically proficient robots risk rejection or suboptimal use in real-world settings.
\vspace{-4.5mm}
\begin{figure}[h!] % "h!" means place it roughly here
    \centering
    \includegraphics[width=\linewidth]{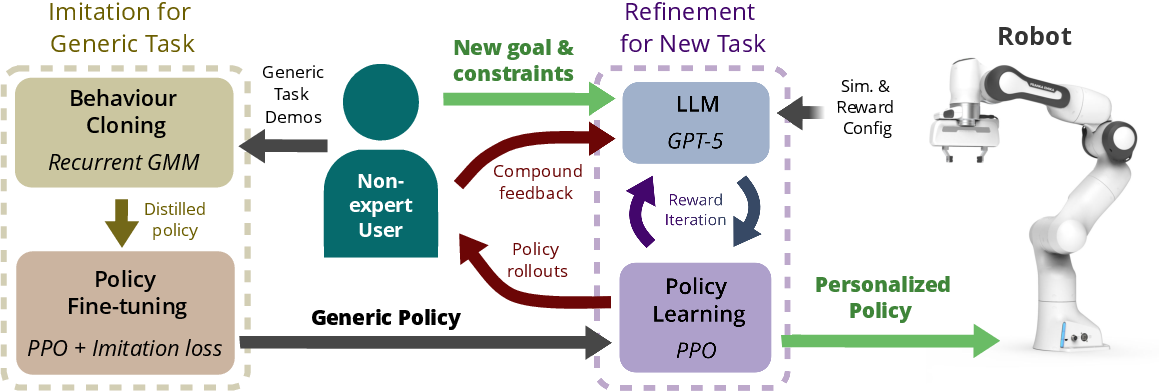}
    \caption{PRISM pipeline overview. First a (non-expert) user provides demonstrations on a generic task to generate a generic policy from imitation (left). Under a new task defined by the user in natural-language, this policy is iteratively refined using an LLM that generates reward candidates and improves them according to user feedback (right).}
    \label{fig:PipelineOverview} % (optional) for referencing later
\end{figure}
\vspace{-10mm}

This paper introduces an instruction-conditioned refinement framework that initializes policies from teleoperated demonstrations and then refines those policies with RL guided by natural-language instructions and iterative human corrective prompts (detailed in \textit{Figure}~\ref{fig:PipelineOverview}). The refinement process is designed to accommodate two classes of deployment shifts: in goal specification e.g. altered target poses requiring new trajectories, and additional manipulation constraints e.g. maintaining an upright vessel while transporting liquid. Natural-language instructions provide an interpretable specification of these modified objectives, while human-in-the-loop feedback ensures targeted correction during refinement. Together, these mechanisms enable PRISM to (i) reduce sample complexity relative to scratch policies that don't rely on human feedback, (ii) preserves the advantageous priors learned from demonstrations to avoid reward-exploiting behaviors, and (iii) provides a practical pipeline for rapid task refinement and personalization. PRISM is put to the test on a representative manipulation task, with results benchmarked against IL-only, RL-only, and instruction-guided refinement methods.

%The remainder of the paper details the data collection protocol, the architecture for instruction-conditioned refinement, the human-in-the-loop prompting process, and an empirical evaluation that compares IL-only, RL-only, and instruction-guided refinement baselines for a task that introduces goal and constraint shifts. 

%
\section{Related Work}

\paragraph{\textbf{Imitation Learning (IL).}} IL trains policies by mimicking human-guided demonstrations and is particularly useful when rewards are hard to specify~\cite{billard16}. It can achieve reasonable performance from relatively few samples, but suffers from limited generalization: small deviations from the training distribution often lead to compounding errors or failure~\cite{codevilla_end_2018}. 
In practice, behavior cloning~(BC) tends to saturate in performance and is brittle under distribution shifts, especially for long-horizon or precision tasks~\cite{rajeswaran17}. Although IL provides a data-efficient initialization, it cannot reliably recover from unseen errors; modern work therefore augments IL with data augmentation or corrective learning, but dependence on demonstration quality and coverage remains a fundamental limitation \cite{mandlekar21}.

\vspace{-1mm}

\paragraph{\textbf{Reinforcement Learning (RL).}} RL learns through trial-and-error by maximizing reward signals, discovering adaptive and robust strategies beyond those demonstrated by humans~\cite{kaelbling96}.  
It has achieved strong results in simulated domains such as locomotion and manipulation, where massive data collection is feasible~\cite{tang24}. However, RL remains sample‐inefficient and difficult to apply in the real world due to safety, time, and reward design constraints~\cite{daly19}. These challenges motivate the integration of structured guidance or demonstrations to accelerate learning. Recent advances in instruction-conditioned and language-guided RL aim to reduce the need for manual reward tuning. Systems like Eureka leverage large language models (LLMs) to automatically generate and refine reward functions from natural-language task descriptions~\cite{ma_eureka_2023_2}. Complementary research explores human-in-the-loop RL, where users provide qualitative feedback or preferences during training to guide policy improvement. These language and feedback driven approaches enhance RL sample efficiency and alignment with human intent by transforming high-level instructions into meaningful learning signals~\cite{yu24,yuann24}.

\vspace{-1mm}

\paragraph{\textbf{IL \& RL.}}  
Contemporary methods commonly combine imitation learning (IL) and reinforcement learning (RL) to leverage demonstrations for sample-efficient initialization and RL for robustness and adaptation~\cite{rajeswaran17,shenfeld23,nair17,torne_reconciling_2024_2}.  Examples include DAPG, which augments policy-gradient updates with a demonstration loss to retain demonstrations priors~\cite{rajeswaran17}, residual-RL schemes that learn corrective controllers on top of a frozen IL base~\cite{ankile_imitation_2024_2}, and recent adapter-based refinements such as Policy Decorator that train lightweight online modules to adjust pretrained policies~\cite{yuan_policy_2025_2}.  These hybrids improve stability and sample efficiency for task-level adaptation but typically rely on engineered rewards, assume known task shifts, and do not provide an interpretable, instruction-driven personalization interface.
In contrast, the present work targets instruction-conditioned personalization: PRISM preserves imitation priors during RL via behavior-matching regularization and uses a language-to-reward loop through Eureka~\cite{ma_eureka_2023_2} with sparse human feedback to specify and refine task objectives without manual reward engineering. This enables rapid, user-directed adaptation to changes in goals and constraints while avoiding retraining from scratch.

\vspace{-1mm}

\section{PRISM's Pipeline}\label{SystemOverview}

PRISM is organized as a modular pipeline that, from a set of teleoperated demonstrations, uses free-form instructions in natural-language to generate a refined and deployable manipulation policy. The main components and data flows are grouped in three principal parts.

\paragraph{\textbf{Data collection.}} Task executions are acquired from teleoperation, allowing a human operator to directly control the manipulator’s end-effector. During each session, multimodal observations, including proprioceptive data, object states, and environment metadata, are captured alongside the corresponding control actions.

\paragraph{\textbf{Imitation Learning.}} Recorded demonstrations are distilled into an initial imitation policy using \textit{Robomimic}~\cite{mandlekar21}. This base policy serves as a behavioral prior for downstream refinement.

\paragraph{\textbf{Reinforcement Learning refinement.}}
The imitation policy is adapted via reinforcement learning guided by natural-language instructions that specify task modifications. 
In line with the overall PRISM framework, the refinement stage is designed to accommodate two representative types of task variation:
\begin{itemize}
    \item \textbf{Manipulation constraints:} additional requirements imposed during execution (e.g., transporting a filled glass while keeping it upright).
    \item \textbf{Goal specification:} shifts in the target pose or placement that require adaptation of the end-effector trajectory.
\end{itemize}
These variation types are expressed through natural-language prompts that are parsed by the \textit{Eureka} module into structured optimization objectives, effectively automating the reward engineering process and steering the RL-based refinement toward the intended behavioral adjustments. 

%Additionally, while the framework inherently supports adaptation to variations in object dynamics, a customary capability of RL, such scenarios are left for future experimental validation.

%
\subsection{Data collection}\label{meth-data-collection}
%

% Expert demonstrations are acquired through a virtual reality (VR) setup that enables immersive teleoperation of the robot manipulator. The operator wears an \textit{HTC Vive Pro 2} headset and uses its hand controllers to control the end-effector’s position, orientation, and gripper state within the simulated environment. This configuration provides intuitive kinesthetic feedback while ensuring accurate pose tracking for fine-grained manipulation tasks.

% Within \textit{IsaacSim}, each demonstration is automatically labeled as successful or unsuccessful according to predefined task completion and failure criteria. The user can also manually reset a trial using the VR controller interface whenever an attempt does not meet quality standards. This procedure ensures a clean, fully labeled dataset and avoids the challenges typically associated with real-world teleoperation, such as calibration drift or embodiment mismatches, thereby eliminating many of the sources of the real-to-sim gap.

% Additionally, demonstrations are automatically segmented into subtasks (e.g., reaching, grasping, placing) to facilitate later data augmentation and modular policy training. These annotations, handled through the \textit{IsaacLab Robomimic} library, enable consistent extraction of relevant motion primitives to support the imitation learning phase. The final dataset includes complete ground-truth trajectories with state, action, and success metadata, suitable for training and benchmarking.

Demonstrations are collected by teleoperating the manipulator and represented as trajectories, $
\tau = \{(s_t, a_t)\}_{t=0}^{T}, \quad \tau \in \mathcal{T}$, where $s_t \in \mathcal{S}$ is the state of the robot and environment at time step $t$, and $a_t \in \mathcal{A}$ is the demo action. They are automatically labeled with binary task success indicators $y_{\text{succ}} \in \{0, 1\}$ based on predefined predicates evaluated on the final state $s_T$ (e.g., object within target zone, upright constraint satisfied). Formally, we define a labeling function, $L_{\text{succ}}(\tau) = \mathbb{1}\{\phi(s_T) \geq \delta\}$, where $\phi(s_T)$ is a task-specific success metric and $\delta$ is a fixed threshold. The operator can manually reset and discard low-quality demonstrations in real time, ensuring a clean dataset.

To support data augmentation, each trajectory $\tau$ is segmented into $K$ semantically meaningful primitives, corresponding to high-level phases such as \textit{reaching}, \textit{grasping}, \textit{transport}, and \textit{placing}. Segment boundaries are computed via rule-based heuristics using object-relative distances, contact conditions, and gripper state transitions, entirely implemented through the IsaacLab-Robomimic toolkit.

The final dataset is denoted by $\mathcal{D} = \{(\tau_i, y_i, \mathcal{P}(\tau_i))\}_{i=1}^{N}$, where $N$ is the number of demo trajectories. Each entry includes full state-action sequences, task success labels, and primitive-level segment annotations. This dataset serves as the foundation for both imitation learning and subsequent refinement.
\subsection{Imitation Learning}\label{meth-IL}
%

% The collected and augmented demonstrations are used to train an imitation learning (IL) policy through a standard behavior cloning (BC) approach. Training is conducted using the \textit{Robomimic} framework, which provides standardized implementations of BC and other IL algorithms. 

% The resulting IL policy serves as a behavioral prior capable of reproducing the demonstrated manipulation task under conditions similar to those seen during data collection. It provides stable initial performance and sample-efficient initialization for the subsequent reinforcement learning refinement stage. 

We use the demonstrations~(see Section~\ref{meth-data-collection}) to train an initial policy via behavioral cloning using a gaussian mixture model with a recurrent neural network (BC-GMM-RNN), a standard imitation learning approach. Each trajectory $\tau_i$ is treated as a supervised sequence of state-action pairs, where the goal is to regress the action from observed states.

We parameterize the policy as a recurrent conditional distribution $\pi_\theta(a \mid s)$ with parameters $\theta$, and optimize via empirical risk minimization:
\begin{equation} \label{eq:bc_gmm_rnn}
\min_{\theta}\; \mathcal{L}(\theta)
= \frac{1}{|\mathcal{D}|}\sum_{i=1}^{N}\sum_{t=0}^{T_i}
\Big[-\log p_{\texttt{GMM}}\big(a_{t}^{i}\mid h_{i,t};\theta\big)\Big]
\end{equation}
where the per-step loss is the negative log-likelihood of the demo action under the RNN-conditioned Gaussian mixture model, and $h_{i,t}$ is the RNN hidden state encoding the history up to time $t$. This objective assumes access to demo action labels drawn from a supervisor policy $\pi_E$ and seeks to minimize the expected deviation under the empirical state distribution induced by the demonstrations.

The resulting policy $\pi_{\theta_{\text{BC}}}$ is trained using the Robomimic framework, which standardizes data loading, loss computation, and batch-wise optimization. The BC policy provides a task-competent behavioral prior that performs reliably under conditions similar to those observed during demonstration. In our pipeline, it also serves as a sample-efficient initialization for downstream reinforcement learning, accelerating convergence and guiding safe exploration.

\subsection{Reinforcement Learning Refinement} \label{refinement}
%

% The imitation policy obtained in the previous stage~(see Section~\ref{meth-IL}) is used as an initialization or behavioral prior for reinforcement learning (RL), enabling the agent to adapt to new task variations introduced through natural-language prompts. These task modifications, detailed in \textit{Section}~\ref{SystemOverview}, \textcolor{red}{include changes in object dynamics, goal configurations, and manipulation constraints.}

To adapt to task variations and personalization objectives, we refine the imitation policy $\pi_{\theta_{\text{BC}}}$ using reinforcement learning. The policy is re-parameterized as $\pi_\theta(a \mid o)$, where $o$ is the observation (e.g., RGB image, proprioceptive state), and is optimized via Proximal Policy Optimization (PPO)~\cite{schulman_proximal_2017_2}. The refinement process comprises two stages: task adaptation and personalization.

% The RL process is divided into two parts, an initial refinement process, and a secondary personalization process. Both of them employ the  \textit{Proximal Policy Optimization} (PPO) algorithm~\cite{schulman_proximal_2017_2}, but with some modifications:
\paragraph{\textbf{Refinement Phase.}} A behavior-matching regularization term is incorporated into the PPO objective (\textit{Eq.}~\ref{eq:Il_loss}) to encourage the refined policy to remain close to the actions of the IL baseline when observing similar states:

% \begin{equation} \label{eq:Il_loss}
% \mathcal{L}_{\mathrm{IL}}(\theta)
% = -\gamma \sum_{(a_i,o_i)\in D^*}^N \log \pi_{\theta}\big(a_i \mid o_i\big)
% \end{equation}

%consider to also report how it looks the combined equation -although it is maybe so simple, so I am doubting-, also maybe using expectations is more elegant?-:

\begin{equation} \label{eq:Il_loss}
\mathcal{L}_{\text{RL}}(\theta) = \mathbb{E}_{(s_t, a_t) \sim \pi_\theta} \left[ \mathcal{L}_{\text{PPO}}(\theta) \right] - \gamma \cdot \mathbb{E}_{(o_i, a_i) \sim \mathcal{D}^*} \left[ \log \pi_\theta(a_i \mid o_i) \right]
\end{equation}

\noindent where $\gamma$ is a scalar coefficient that controls the strength of the behavior cloning prior. This regularization encourages the refined policy to remain close to $\pi_{\theta_{\text{BC}}}$ on demonstrations-distribution observations, improving early sample efficiency and mitigating policy drift.
    % \item \textbf{\textit{Refinement Phase.}} A behavior-matching regularization term is incorporated into the PPO objective (\textit{Eq.}~\ref{eq:Il_loss}) to encourage the refined policy to remain close to the actions of the IL baseline when observing similar states. This is achieved by sampling from a dedicated demonstration replay buffer $D^*$ containing $N$ expert observation–action pairs. The contribution of this regularization is weighted by a scalar coefficient $\gamma$, enabling controlled influence over policy updates during refinement. This mechanism stabilizes early learning, mitigates policy drift away from demonstrated behavior, and improves sample efficiency while the agent adapts to the modified task objectives. \textcolor{red}{29.42 min for 98\% accuracy, 5.000 environments inside IsaacSim}

\paragraph{\textbf{Personalization Phase.}}  
After task-level refinement, we adapt the policy to user-specific variations expressed through natural language prompts. Let $\mathcal{I}$ denote the space of language instructions, and let \mbox{$f_{\text{NL2R}}: \mathcal{I} \!\rightarrow\! \mathcal{R}$} be a structured mapping from an instruction $\iota \in \mathcal{I}$ to a reward function. This function is implemented via IsaacLab’s adaptation of the Eureka framework\footnote{https://github.com/isaac-sim/IsaacLabEureka}, which compiles prompts into shaped reward components.

At training time, the reward used for policy optimization is expressed as $r_t = r_t^{\text{base}} + r_t^{\text{aux}} + r_t^{\text{pers.}}$,
where $r_t^{\text{base}}$ captures core task objectives (e.g., goal reaching), $r_t^{\text{aux}}$ may include time penalties or regularization terms to encourage stability and efficiency, and $r_t^{\text{pers.}} = f_{\text{NL2R}}(\iota, s_t, a_t)$ encodes prompt-specific shaping.

To improve alignment with the user's intent, prompts $\iota$ are iteratively updated via a hybrid mechanism. Automated prompts are generated by a policy $\pi_{\text{auto}}(\pi_\theta)$ based on task-specific evaluation criteria (e.g., signals extracted from the simulated environment such as observed state transitions, task success/failure statistics, reward value trajectories, constraint violations, and qualitative failure modes identified during rollouts), while human-in-the-loop prompts are introduced at predefined checkpoints. This hybrid prompting loop enables scalable yet targeted adaptation to evolving task demands or user feedback, without requiring continuous supervision.

\begin{figure}[h!] % "h!" means place it roughly here
    \centering
    \includegraphics[width=\linewidth]{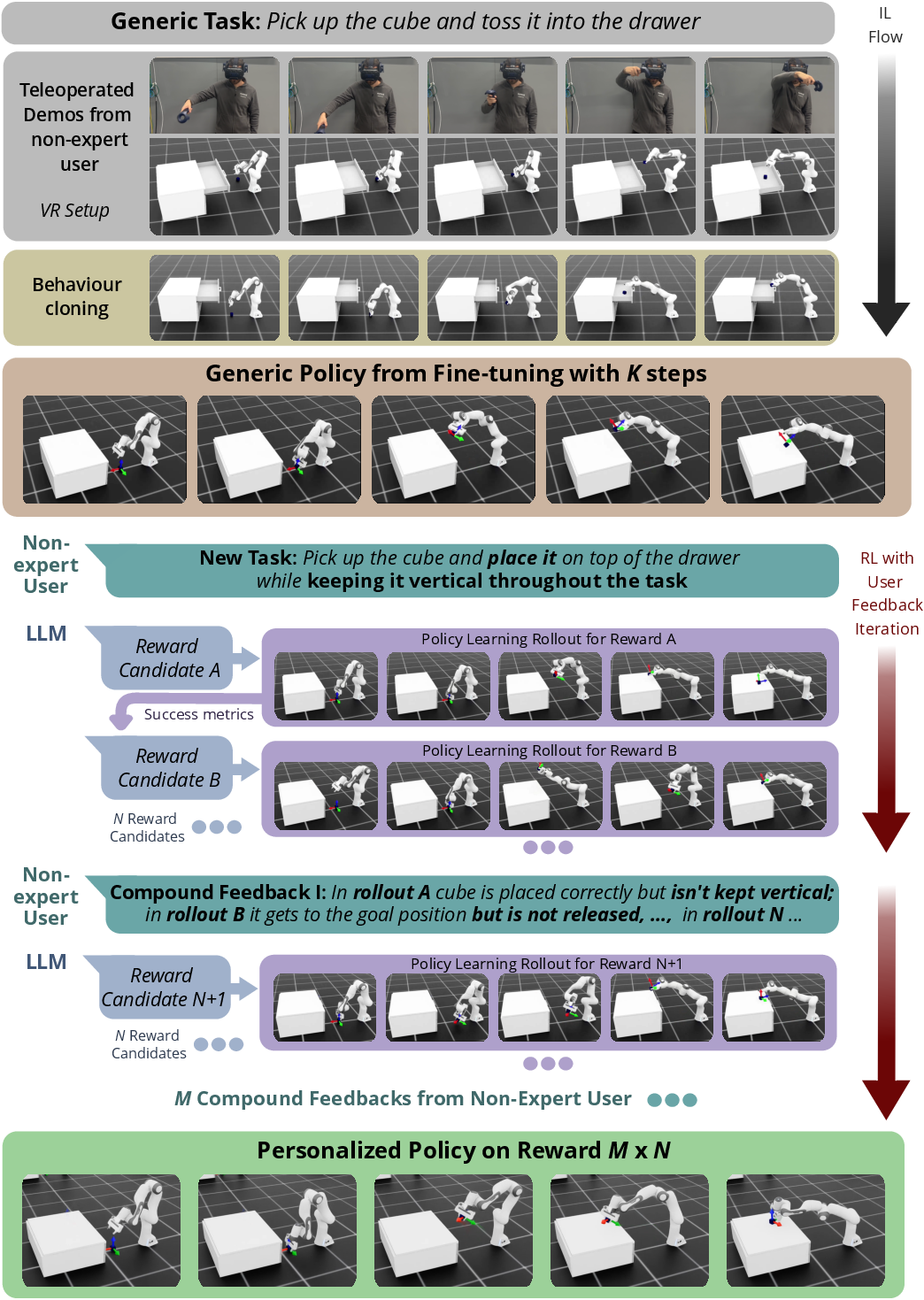}
    \caption{Execution of PRISM pipeline exemplified through evaluation experiment.}
    \label{fig:PipelineOverview} % (optional) for referencing later
\end{figure}

\vspace{-3mm}
\section{Experimental Evaluation}
\vspace{-2.5mm}
PRISM is evaluated on a manipulation task involving the personalization of a general pick-and-toss policy into a constrained pick-and-place policy. The complete PRISM pipeline has been depicted for this experiment in \textit{Figure}~\ref{fig:PipelineOverview}.

Therefore, the objective of the general policy is to pick a cube and toss it into a cupboard, regardless of the intermediate motion trajectory or final object orientation. The demonstrations are collected using a virtual reality (VR) teleoperation setup inside \textit{Omniverse}'s \textit{IsaacSim} simulator~\cite{isaacsim25}. A non-expert human operator wearing an HTC Vive Pro 2 headset and controllers teleoperates robot's end-effector pose and gripper state, allowing intuitive and precise execution of complex manipulation tasks. Observation space includes the cube and goal poses relative to the manipulator’s end-effector, the manipulator’s joint positions and velocities, and the previous control action. Actions are defined as relative task-space deltas applied to the end-effector pose and the gripper state.

The collected dataset, consisting of 50 demonstrations, is used to train an IL baseline for 500 epochs (31 min) and yields a task success rate of 21.2\%\footnote{All training was done on a PC with an NVIDIA GeForce RTX 3080 Ti.}, frequently failing when the environment deviates from the demonstrated trajectories. To improve its robustness, generated policy is finetuned using RL following the method described in \textit{Section}~\ref{refinement}. The agent is trained with a task reward that solely evaluates goal completion, leveraging 5,000 parallel environments. This fine-tuning phase converges in 1000 steps, adding up to 5M total steps (5,000 environments x 1,000 steps), achieving a task success rate of 98\% after 29.4 minutes, outperforming the IL baseline.

PRISM’s personalization capabilities are assessed by adapting the refined policy to a new task that (i) modifies the goal state by requiring the cube to be placed precisely on a tabletop instead of tossed, and (ii) introduces an additional manipulation constraint enforcing the cube’s verticality throughout the task. Personalization begins from the final checkpoint of the fine-tuned policy, guided by a natural-language prompt describing the new task. The instruction-to-reward refinement process uses the \textit{GPT-5} model and iterates automatically, with one human feedback prompt introduced after every $N=5$ automated \textit{Eureka} iterations. This loop is executed $M=2$ cycles, after which the personalized policy achieves a final success rate of 96.8\%.

\begin{figure}[h!] % "h!" means place it roughly here
    \centering
    \includegraphics[width=\linewidth]{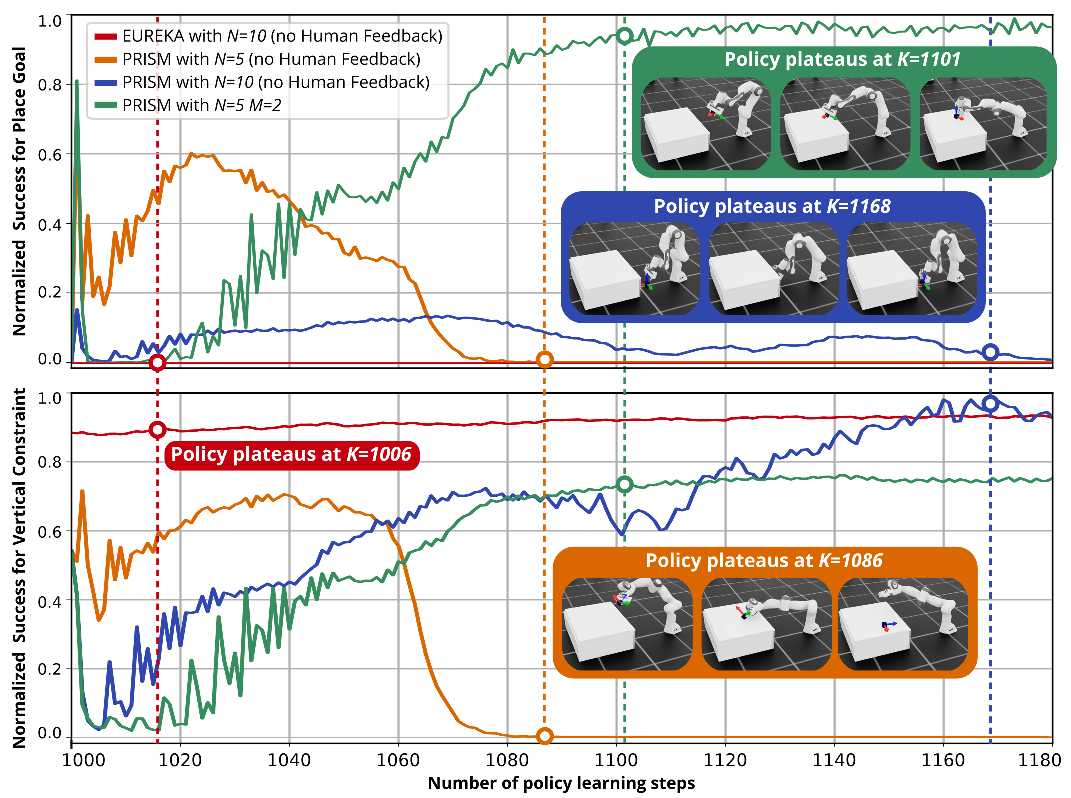}
    \caption{Evolution of success metrics for place goal and vertical constraints across four methods. Policies are deemed learned when the total episode reward remains stable for a sustained interval of training steps, marked on both plots with colored dots. For illustration, representative post-training trajectories for PRISM are shown; the EUREKA (RL-only) baseline is excluded from the plot as it stayed to an unsuccessful idle policy.}
    \label{fig:Graphs} % (optional) for referencing later
    \vspace{-4.5mm}
\end{figure}

The influence of human feedback frequency is evaluated by comparing the hybrid (automated + human) and fully automated personalization strategies. As illustrated in \textit{Figure}~\ref{fig:Graphs}, the inclusion of sparse, well-timed human feedback notably accelerates convergence and enhances final task performance, demonstrating the efficiency of limited yet targeted human intervention. Conversely, the fully automated approach exhibits slower reward adaptation and greater variability in task success. For comparison, the figure also includes the training curve of a baseline using only the \textit{Eureka} reward generation process-without initialization from the imitation policy—which fails to achieve the task after ten iterations of 1500 steps each, underscoring the importance of imitation priors for effective policy refinement.

\vspace{-3mm}
\section{Conclusions and Future Work}
\vspace{-3mm}
This work introduced \textbf{PRISM} -\emph{Personalized Refinement of Imitation Skills for Manipulation via Human Instructions}- a pipeline that integrates IL, reinforcement-based refinement, and instruction-conditioned personalization. Policies are initialized via BC from teleoperated demonstrations and subsequently refined with language-guided RL augmented by sparse human corrective feedback (PRISM overview video \ref{footnotevideo}). The combined approach preserves the stability and data efficiency of imitation while providing the adaptability required for dynamic, user-specific manipulation tasks. Moreover, by allowing end-users to guide refinement through natural-language instructions, PRISM reduces dependence on expert-crafted reward functions and facilitates the configuration of personalized manipulation behaviors in real environments. 

Experimental results indicate that PRISM achieves 96.8\% success-rate for a concrete task adaptation scenario with a total duration of 4 hours, being the only method to complete the proposed task with the stated variations in that time. The integration of human instructions within the refinement loop enables scalable personalization without the need for continuous supervision, while maintaining interpretability and efficient reward specification, illustrating its practical value for deployable, user-adaptive robotic systems.

Experiments were conducted entirely in simulation, leaving real-world dynamics, perception noise, and hardware constraints unvalidated. The current personalization protocol depends on occasional human feedback and explicit success criteria, which may limit scalability across diverse users, task, and longer interaction horizons. Additionally, the refinement stage requires a suitable simulator and episodic resets, restricting applicability to environments where such assumptions hold. These limitations motivate several directions for future work: closing the sim-to-real gap; systematically evaluating scalability across varied users, task families, and long-term interactions; and advancing refinement strategies through implicit preference inference, continual personalization mechanisms, and safer online adaptation on real robotic hardware.

\vspace{-4mm}

\section{Acknowledgments}
\vspace{-3mm}
This work was financially supported by the European Commission’s Horizon Europe Framework Program through the IntelliMan project under Grant Agreement No. 101070136; and by the Catalan Government through ACCIO-Eurecat (Project Traça-SESAM) funding grant. Views and opinions expressed are however those of the author(s) only and do not necessarily reflect those of the funding institutions. The funding institutions cannot be held responsible for them.
\vspace{-4mm}
% %
% \section{Appendix}
% %
% \textcolor{blue}{Hyperparameters, architecture details, complete prompts, additional plots, dataset/demos specification, code release instructions}

\bibliographystyle{plain}
\bibliography{other-references,zotero-references}

\end{document}